A Robust Classification Method using Hybrid Word Embedding

for Early Diagnosis of Alzheimer's Disease

Yangyang Li

Massachusetts Institute of Technology

Cambridge MA, United States

# ABSTRACT

Early detection of Alzheimer's Disease (AD) is greatly beneficial to AD patients, leading to early treatments that lessen symptoms and alleviating financial burden of health care. As one of the leading signs of AD, language capability changes can be used for early diagnosis of AD. In this paper, I develop a robust classification method using hybrid word embedding and fine-tuned hyperparameters to achieve state-of-the-art accuracy in the early detection of AD. Specifically, we create a hybrid word embedding based on word vectors from Doc2Vec and ELMo to obtain perplexity scores of the sentences. The scores identify whether a sentence is fluent or not and capture semantic context of the sentences. I enrich the word embedding by adding linguistic features to analyze syntax and semantics. Further, we input an embedded feature vector into logistic regression and fine tune hyperparameters throughout the pipeline. By tuning hyperparameters of the machine learning pipeline (e.g., model regularization parameter, learning rate and vector size of Doc2vec, and vector size of ELMo), I achieve 91% classification accuracy and an Area Under the Curve (AUC) of 97% in distinguishing early AD from healthy subjects. Based on my knowledge, my model with 91% accuracy and 97% AUC outperforms the best existing NLP model for AD diagnosis with an accuracy of 88% [32]. I study the model stability through repeated experiments and find that the model is stable even though the training data is split randomly (standard deviation of accuracy = 0.0403; standard deviation of AUC = 0.0174). This affirms our proposed method is accurate and stable. This model can be used as a large-scale screening method for AD, as well as a complementary examination for doctors to detect AD.

# 1. Introduction

Alzheimer's Disease (AD) is the first leading cause of dementia that involves a decline of memory, impairment of language ability, and other general cognitive skills [1]. Alzheimer's Disease is the reason for 60% to 80% of dementia and the sixth major cause of mortality in the United State. It is also a tremendous burden to this country. In 2012, 200 billion dollars' care expense was spent on it. Since there is no cure for the disease, early detection is crucial.

Early diagnosis is very challenging because many early signs of AD are similar to normal aging. Moreover, current diagnosis of AD is time-consuming, expensive, and may have harmful side-effects. The most widely used methods include Mental Status Tests like MMSE (Mini-Mental State Examination) and medical imaging-based such as CT, MRI, and PET. The interview based MMSE test has an early detection accuracy of 87% [19]. However, detection accuracy highly depends on the clinician's level of experience if mental status tests are administered manually. Medical imaging methods typically have accuracy ranging from 80% to 95%. For Medical Imaging, the cost for laboratory tests and so on is high (a PET scan costs 3000-5000 dollars) [16]. Additionally, some medical imaging involves exposure to radioactivity, such as PET. This may cause harmful side-effects to the human body.

The research presented in the paper uses a novel method based on machine learning and natural language processing (NLP). Using NLP, my model requires only one task before offering the diagnosis: a spoken description of a picture. The simple diagnosis has minimum effect on subjects and has an accuracy of 91%. At present, the accuracy of other best NLP models that use text information to detect AD is close to that of doctors. However, there still remains a gap. Thus, using innovations in hybrid word embedding and hyperparameter tuning, I propose a new model that exceeds the doctor's detection accuracy. I also prove the model to be stable. My model has an accuracy of 91%, an AUC of 97%, and outperforms the best NLP model for AD diagnosis of Sarawgi, Utkarsh, et al. with an accuracy of 88% [2]. I use hybrid word embedding to get a comprehensive representation of the semantic context. This embedding combines word vectors from Doc2vec and ELMo which provides a deep contextualized word representation. Furthermore, the model is incorporated in an online application for practical use. This web-based software, AD Scanner, can significantly improve the efficiency of Alzheimer's diagnosis and offers a complementary examination for doctors to detect AD. The user is only required to speak about one picture for one minute, and a detailed result is generated within seconds.

I optimize the hyperparameters in feature engineering steps and the classifier using cross-validation. I evaluate models by studying stability in repeating my experiments and examining ROC curves. I also repeat the experiment 1000 times. The results show that my models are stable, and the standard deviation of accuracy is less than 5% of the average accuracy for every model.

I compare the proposed method with three baseline methods. I use only word count in the first model and received a baseline accuracy of 81% and an AUC of 88%. In the third model, I include text embedding from Doc2vec as a semantic feature and boost the accuracy to 89% and the AUC to 95%. In the fourth model, I achieve the highest accuracy of 91% and an AUC of 97% by using hybrid word embedding with linguistic features and word count. My methods only use a basic classification algorithm logistic regression, but I have combined multiple features and tuned hyperparameter across the entire machine learning pipeline. The best model achieves a relatively high accuracy of 91%.

The main contributions of my paper can be summarized as follows:
1. I have developed a new classification method using hybrid word embedding for early detection of AD.
2. I have enriched the word embedding feature with linguistic features to capture additional information about the response of the interviewee.
3. I have tuned hyperparameters for both the feature engineering step and the classification step using random search and cross-validation.
4. I have demonstrated stability of the classification model by through repeated experiments.
5. I have developed an online application for practical use of my proposed model.

## 2. Related Work

Much of the previous work has been done to predict AD given participants' language ability. Sarawgi, Utkarsh, et al. used artificial neuron network with temporal characteristics to detect AD and has an accuracy of 88% [2]. Gaspers et al. used the machine learning model of lexical, sublexical and temporal features to minimize the numbers of required classification tasks [23]. Methods used include support vector machine [6], and a linguistic model that has been used by Turin University Linguistic Environment (TULE) [7]. Different environments from TULE were taken into consideration when Boye, Tran and Grabar used ecological conversation situation [8]. Khodabakhsh et al. used a different source of data from [8] for preliminary analysis of automated AD early detection through features derived from conversation transcriptions [9]. Even more complicated models have also been used by Karlekar, Niu and Bansal. They used Convolutional Neural Network in combination with Long-Short Term Memory model and achieved great results [10]. The validity of using complicated syntactic features to classify MCI (the precursor of AD) was demonstrated in [3], a study that distinguished MCI and healthy elderly using spoken language features. This study found seven statistically significant linguistic features, and 86.1% AUC was showed on the combination of those linguistic features and multiple test scores. Rudzicz et al. used Carolina Conservation Collection and DementiaBank to extract the most indicative lexical/syntactic and acoustic features of AD [18]. Fraser also derived data from DementiaBank. He conducted an exploratory factor analysis on the extracted features of language transcripts and obtained state-of-art classification accuracies over 81% [5]. Meanwhile, Rentoumi analyzed the language samples of a longitudinal clinical study from patients and found that syntactic and lexical features yielded low discrimination scores while word occurrences and frequencies offered the highest accuracy [12]. Recent studies even derived data from Tweeter [4] to train the classifier to discriminate AD patients on social medias [4].

The current work in early diagnosis of AD still has limitations. First of all, the detection procedures in other pre-existing works usually include multiple tasks. For example, MMSE, General Practitioner assessment of Cognition (GPCOG), Montreal Cognitive Assessment (MoCA), and Mini-Cog have participants take numerous supervised conversation tests [7]. However, the only task my model requires is a description of the picture "Boston Cookie Theft". Secondly, most research use basic feature engineering approaches, which limits the accuracy of the classification model. Thirdly, machine learning models can be complicated and have low interpretability. For example, a deep learning model requires a time-consuming model training process, a complicated comprehensive model tuning, and it could have low interpretability.

## 3. Data Description and Preprocessing

### 3.1. Dataset

Dataset used in this paper is from DementiaBank. It is the largest dataset publicly available for cognitive impairment detection from human speech developed by the School of Medicine at the University of Pittsburgh [17]. Specifically, I use English transcripts of the Pitt corpus [24]. The data contains transcripts and audio recordings of a group of probable AD experiment participants and a group of control participants that do not have AD. Researchers recorded interviews where participants were asked to perform tasks that revealed their language capabilities. The task uses the "Boston Cookie Theft" picture and asks the participants to describe what they see in the picture.

The demographic data and test results from the above protocol are collected on the transcripts. The dataset contains age, gender, and MMSE scores of the participants, and the raw data contains information about the number of pauses, unintelligible objects, repetitions, etc. detected during the conversation. The sample consists of 300 randomly chosen recording interviews of participants aged between 40 and 80 (150 males and 150 females, the average age is 60 years old, SD = 10), including 104 healthy control , 208 dementia patients, and 85 people with unknown diagnosis.

## 3.2. Data Preprocessing

Interview recordings of DementiaBank are stored in files with CHAT format [22] which represents different linguistic components using special characters. For example, unintelligible words are transcribed by 'xxx' when the interviewer cannot hear or understand what the participant is saying. Here is an instance of the use of the 'xxx' symbol:

**\*GAB:      I want xxx .**
**\*SAM:      what do you want ?**
**\*GAB:      I said I want xxx .**

Repetition, defined as Retracing Without Correction, is transcribed in the transcript by enclosing the speech that is repeated within angle brackets (< >), followed by square brackets ([/]) with an inside slash mark, as in this example:

**\*DAV:      <but but but> [/] but (.) it's a cat.**

In my machine learning pipeline, I start with data cleaning by removing punctuations. Then I replace "\ -", "\ /", etc. with space, and delete punctuations such as ".", ":", "?", ";", etc.

## 4. Methods

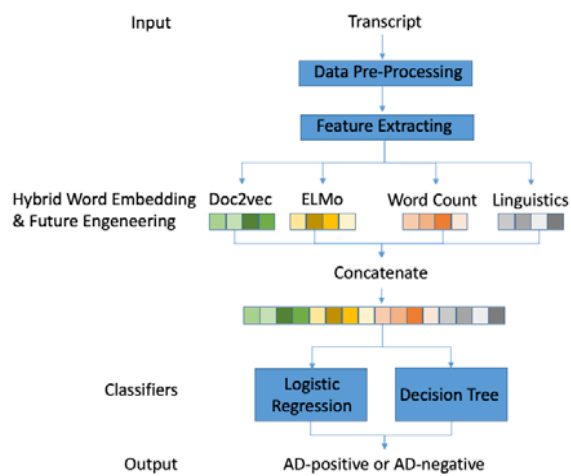

 Figure 1: Methods

## 4.1. Hybrid Word Embedding

I further develop hybrid word embedding features to capture semantic context by combining word vectors from Doc2vec and ELMo.

Doc2vec [20] uses unsupervised learning to learn fixed-length feature representation from documents. It is an efficient, widely-used algorithm that can encode semantic context of a document in a vector. Unlike sequential models such as RNN, in which the word sequence is generated as sentence vectors, the Doc2vec sentence vectors are independent of word order. Given $M$ words $t_1, t_2, ..., t_M$, Doc2Vec learns such a set of word vectors $\{v_t\} \subset \mathbb{R}^n$ for each word and a set of document vectors $\{v_d\} \subset \mathbb{R}^n$ for each document. Each embedding is close to the embedding of its "context". The context of word $t_i$ is a set of words that appears nearly $t_i$ in corpus, e.g., $k$ words around $t_i$, and the documents which contain word $t_i$. Formally, context of $t_i$ is

$C_i = \{v_j \mid t_j \text{ appears nearly around } t_i \text{ within } k \text{ steps}\} \cup \{v_d \mid \text{document } d \text{ contains word } t_i\}$

The goal of Doc2Vec is to maximize the softmax probability between each word vector $v_i$ and the vectors of its context $C_i$:

$$\Pr(v_i|C_i) = \prod_{v_j \in C_i} \Pr(v_i|v_j) = \prod_{v_j \in C_i} \frac{\exp(v_i^\top v_j)}{\sum_{k=1}^{k=M} \exp(v_k^\top v_j)} \quad (1)$$

Iterate eq. (4) over all words $t_i$ and multiply them together to get the whole probability.

$$P = \prod_{i=0}^{i=M} \Pr(v_i|C_i) \quad (2)$$

Then, I minimize the negative logarithm of the whole probability to get the word vectors $\{v_t\}$ and document vectors $\{v_d\}$: $\operatorname{argmax} \sum_{i=1}^{i=M} \sum_{v_j \in C_i} \log \frac{\exp(v_i^\top v_j)}{\sum_{k=1}^{k=M} \exp(v_k^\top v_j)}$.

Another embedding I use is ELMo[21]. ELMo uses bidirectional long short-term memory (LSTM), which allows it to speculate sentence context. Using ELMo with Doc2vec offers statistical features that give a full picture of the document. I also use the logistic regression classification model to efficiently handle a large number of features. Thus, the combined word vectors do not bring much computation cost and improve the prediction accuracy of logistic regression. ELMo uses a forward LSTM model and a backward LSTM model to capture the dependency of a word on preceding and following words. It works out the contextualized embedding by grouping initial embedding and a few hidden states together with concatenation after weighted summation [15].

Elmo is a combination of multi-layer representations of the bidirectional language model (biLM). For each certain token $t_k$, an L-layer bidirectional language model biLM can be represented by 2L+1 vectors:

$$R_k = \left\{x_k^{LM}, \overrightarrow{h}_k^{LM,j}, \overleftarrow{h}_k^{LM,j} \mid j = 1, ..., L\right\} = \left\{h_k^{LM,j} \mid j = 0, ..., L\right\}, \quad (3)$$

where $\left\{h_k^{LMj}\right\} = \left[\overrightarrow{h}_k^{LMj}; \overleftarrow{h}_k^{LMj}\right]$.

In the downstream model, ELMo integrates the output R in Equation 1 of the multi-layer biLM into a vector, $ELMo_k = E(R_k; \theta_e)$. The simplest case is that ELMo only uses the top output, that is $E(R_k) = h_k^{LM,L}$. However, Peters et al. found the best ELMo model is to add the output of all biLM layers to the weights learned by the normalized softmax $s = Softmax(w)$ [21]:

$$E(R_k; w, \gamma) = \gamma \sum_{j=0}^{L} s_j h_k^{LM,j}, \qquad (4)$$

where γ is a scaling factor. If the output of each biLM has a different distribution, the γ is equivalent to using layer normalization for each layer of biLM before weighing.

The vector I derived from doc2vec is denoted as $v_{doc2vec}$. Similarly, I denote the vector derived from ELMo as $v_{ELMo}$.

$$v = (v_{doc2vec}, v_{ELMo}), \qquad (5)$$

**4.2. Feature Engineering**

*4.2.1 Word Count*

I used count vectorizer to retrieve word count from each transcript. Count vectorizer method counts the number of times a token appears in the document and uses this value as its weight [11]. The method evaluates the number of word appearances in the control and dementia group and takes these as input features to a logistic classifier to make a prediction. Linguistic pattern can be learned by BOW transformer and converted to useful information for next step. Using count vectorizer for a randomly selected training set, I pre-analyzed word appearance and plotted the words that have the most different patterns in control group and dementia group.

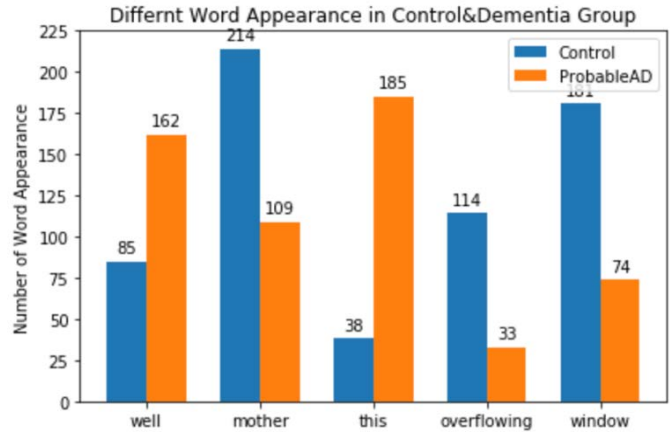

**Figure 2: Different Word Appearance in Control and Dementia Group**

For example, the dementia group tends to use "lady" to describe the woman who is washing dishes (the word "lady" appears 45 times in dementia group while they appear 28 times in the control group).

*4.2.2 Linguistic Characteristics and Demographic Features*

Clinical research shows that there are some linguistic features and demographic features closely related to AD. AD affected-patients usually have varying degrees of language impairment, so I measured word rate, intervention rate, unintelligible word count, and trailing and pause rates of their speech. All linguistic features are normalized by the respective audio lengths and scaled thereafter. AD often begins in people over 65 years of age and is more common among women than men, so I also took demographic features such as age and gender into consideration.

**4.3. Machine Learning Model Development**

I have picked two simple yet normally effective machine learning algorithms for the study: *decision tree* and *logistic regression*.

*4.3.1 Decision Tree*

Decision trees can naturally treat mixtures of numeric and categorical variables and scale well with large data sets. They also manage missing variables effectively. The most important advantage of the decision tree for my model is that decision trees are easily interpretable; in other words, it is possible for a human to understand the reason of the output of the learning algorithm. Thus, it offers an understandable interpretation of the model's predictions so that I can apply it in the real-world applications safely.

Table 1 showed the accuracy of decision tree model in my study. After evaluating the initial baseline models' performance, both without fine tuning, logistic regression shows a better performance than decision tree does. As a result, in the subsequent experiments, I focused on tuning logistic regression.

| Method | Test Accuracy | Parameter |
|---|---|---|
| Decision Tree id3 (gini) + Count Vectorizer + Doc2vec | 56. 06 | gini |
| Decision Tree id3 (entropy) + Count Vectorizer + Doc2vec | 51.46 | entropy |
| Decision Tree + Count Vectorizer + Doc2vec | 65.63 | gini |

**Table 1: Summary Statistics for Test Accuracy using Decision Tree**

*4.3.2 Logistic regression*

Logistic regression is robust because the hessian matrix of the log likelihood function is positive definite which guarantees the existence of a unique minimum. Logistic regression is also computationally efficient, and the combined features do not bring much computation cost but improve the prediction accuracy.

## 5. Experiments and Results

### 5.1. Hyperparameter Tuning

| Hyperparameter | Algorithm | Range |
|---|---|---|
| c (inverse regularization parameter) | Logistic Regression | [0.01, 0.05, 0.25, 0.5, 0.75, 0.99, 10, 1000, 1e6, 1e10 ] |
| vec_size (vector size) | Doc2vec | [40, 100, 200, 500, 750, 1000, 1250, 1500] |
| alpha (learning rate) | Doc2vec | [0.001, 0.025, 0.01, 0.25, 0.35, 0.05, 0.1, 0.5] |
| min-alpha (learning rate) | Doc2vec | [0.00025, 0.0025, 0.005, 0.01, 0.05, 0.1, 0.2, 0.5 ] |

**Table 2: Summary of Hyperparameters Optimized**

Before I train the model, except for inputting the data, features, and labels, I specify the ranges of a set of hyperparameters according to table 2. I experiment with hyperparameter combinations using cross-validation. Each time, I train the model with a selected hyperparameter combination. Then, using accuracy as the evaluation metric, I evaluate the selected hyperparameter set on the validation set. After cross-validation, the best model and the hyperparameter combination are generated. For instance, one hyperparameter I tune in logistic regression is the inverse regularization parameter c as regularization is a constraint that prevents the model from over-fitting or under-fitting.

To train my model, I use cross validation to choose the best hyperparameters in the following steps:

Step 1: Dividing data into training/validation/test set.
Step 2: Determining the range of hyperparameter manually.
Step 3: Searching through the hyperparameter space and finding the best set that gives the highest validation accuracy.
Step 4: Training a model with the best hyperparameters based on step 3.
Step 5: Evaluating the performance of the best model performance on the test set.

Data was divided into three sets. The training set consists of 80% of the data, and the validation and test set both consist of 10% of the data. By repeating experiments, the randomness introduced by randomly splitting data is averaged out.

## 5.2. Area Under Curve

In Machine Learning, performance measurement is an essential task. When it comes to a classification problem, I can count on an AUC - ROC Curve. Area Under Curve (AUC) tells how well the model is classifying between classes. AUC has its vertical axis as the True Positve Rate (TPR) and its horizontal axis as the False Positive Rate (FPR). TRP and FPR are defined as follow: $TPR = \frac{TP}{TP+FN}$ and $FPR = \frac{FP}{TN+FP}$. AUC measures the area under ROC curve from (0,0) to (1,1). Compared to accuracy, AUC is a relatively stable measurement of the model performance when data is imbalanced.

### 5.2.1 Model Accuracy

Logistic regression generates a value between 0 and 1 to indicate the probability of the correct prediction. To calculate the training error, I look at the error at y = 0 and y = 1. Logistic regression uses log loss as its lost function. Although accuracy is generally a good method for evaluating the model performance, confusion matrix fits logistic regression better. Confusion matrix is a square matrix with the numbers of True Positive (TP), True Negative (TN), False Positive (FP) and False Negative (FN). It not only reports the number of wrong predictions but also reports how they are wrong.

As mentioned above, TPR and FPR are particularly useful when the data is imbalanced. The AUC curve is useful for data evaluation in classifying interview transcripts of both healthy and AD groups (although in the dataset the data is not very imbalanced. I use AUC considering the practical use of my model in real life where the morbidity of AD is low). Each data sample is a subject in the DementiaBank dataset. There are two kinds of data points: positive samples and negative samples. Usually positive sample indicates patients, while negative sample indicates healthy person. In my case, one positive sample is an AD patient, and y = 1. One negative sample is a healthy person in the control group, and y =0. Very commonly, negative and positive samples do not each take half of all of the data. When the proportion of healthy people in the data has a large difference with the proportion of AD patients in the data, the data is imbalanced.

The higher the AUC, the better the model is at predicting 0s as 0s and 1s as 1s. A higher AUC value means the classification model has better unbiased performance. An AUC of 1.0000 suggests that the model has perfect performance. In my case, the higher the AUC, the better the model is at distinguishing between patients with disease and without disease.

## 5.3. Model Stability Study

In practice, the model accuracy will be dependent on the test data and many other random factors such as random initialization of the model parameters and the randomness in stochastic gradient descent algorithm. Clearly measuring the variance of model accuracy can reveal model performance and stability in different situations. This is usually achieved by repeating model training and prediction which I have done with the DementiaBank dataset. I evaluated my models by studying stability in repeating experiments and examining ROC curves. I repeated experiments of each pipeline for 1000 times and found out that they are stable even the data is perturbed. According to results, my models are stable and the standard deviation is less than 5% of the accuracy. I will further compare the stability of each machine learning pipeline in Section 5.7.

**5.4. Machine Learning Pipelines**

Four models that incorporate NLP elements are compared in my paper. For all the models, logistic regression acts as the classifier, while word embedding is used before the classifier. Each pipeline takes different feature engineering steps as summarized below:

1. Pipeline 1 uses the statistical features generated by count vectorizer and the classification model logistic regression only.
2. Pipeline 2 adds linguistic characteristics such as number of pauses, unintelligible words, etc. detected in the transcript and demographic features including age and gender.
3. Pipeline 3 adds vector features generated by Doc2Vec to incorporate paragraph information [13].
4. Pipeline 4 adds hybrid word embedding by combining vector features generated by deep learning algorithm ELMo and vector features generated by Doc2vec. ELMo goes a step further and trains a bi-directional LSTM. It comes up with the contextualized embedding through grouping together hidden states (and initial embedding) in a certain way (concatenation followed by weighted summation) [15].

Logistic regression is a widely used classification model. It has a probabilistic output that can be easily interpreted for binary prediction problem.

**5.5. Cross Validation**

Here is an example for cross validation. For every inverse regularization value, I have 10 experiments with different training and validation data. Average accuracy is calculated to be the accuracy given corresponding regularization strength.

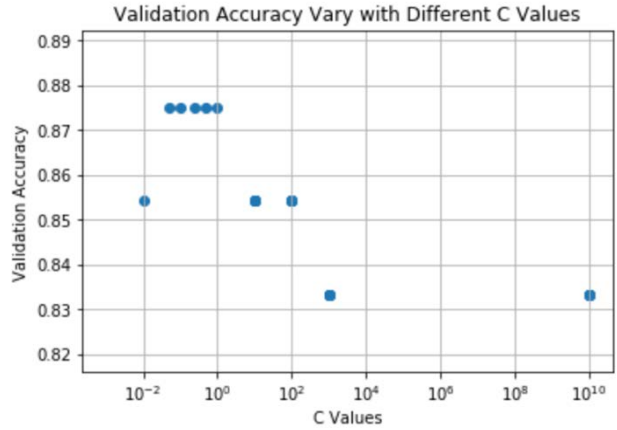

**Figure 3: Validation Accuracy Given Different C Values**

From cross validation results, I see that accuracy peaks when C equals 0.05, 0.25, 0.5, 0.75, and 0.99, but results may vary as the data is perturbed. However, I repeated the experiment multiple times to average out the randomness. This value will be used in subsequent experiment to calculate final results.

### 5.6. Results

I have described the four pipelines in section 5.4. Machine Learning Pipelines. Results of each pipeline include an average accuracy of 1000 experiments, average Test AUC of the experiments, and standard deviations of the 1000 accuracies and AUC values, summarized in Table 3 of this section. Pipeline 1 with word count as NLP elements representation followed with a logistic function as the prediction layer serves as my baseline. The average of 1000 accuracies generated after all data splitting experiments is 0.81: the baseline of all four pipelines. The average test AUC of Pipeline 1 is 0.8897 with a standard deviation 0.0464. Compared to the test AUC values of other three pipelines, the test AUC of pipeline 1 is the lowest, and its standard deviation is the highest. This indicates that, of all four pipelines, pipeline 1 is the most unstable. Below is the distribution of 10 sample AUC values of pipeline 1:

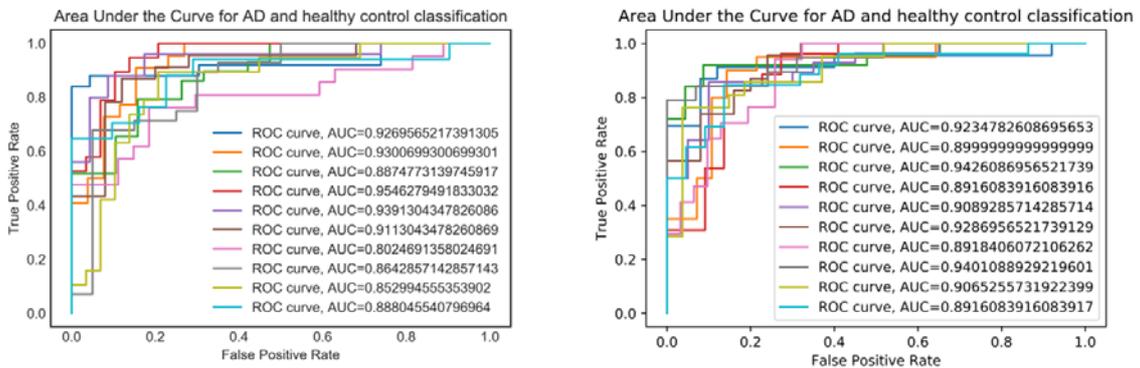

(a)                                   (b)

**Figure 4: ROC and AUC for Pipeline 1 (logistic regression + word count features) on the testing set & ROC and AUC for Pipeline 2 (logistic regression + word count + AD-related features) on the testing set**

In Figure 4 (a), the lines are scattered and vary from each other. This indicates the relatively low stability of the classification performance of pipeline 1.

Pipeline 2 adds AD-related features to word count features. This includes linguistic characteristics extracted from interview transcripts such as number of pauses, unintelligible words, etc. and demographic features age and gender.

Pipeline 4 incorporates word count, AD-related features, and hybrid word embedding transformed vectors into the feature space. This pipeline achieves the best classification results.

Incorporating hybrid word embedding feature representations into pipeline 4 offers the best results. The highest AUC attained was 100%, meaning all cases were correctly classified. The lowest AUC was 95.1% which is still higher than the previous pipelines. It can also be seen from the picture that the variance of AUC distribution decreases as well, indicating that the model is not only more accurate but more robust.

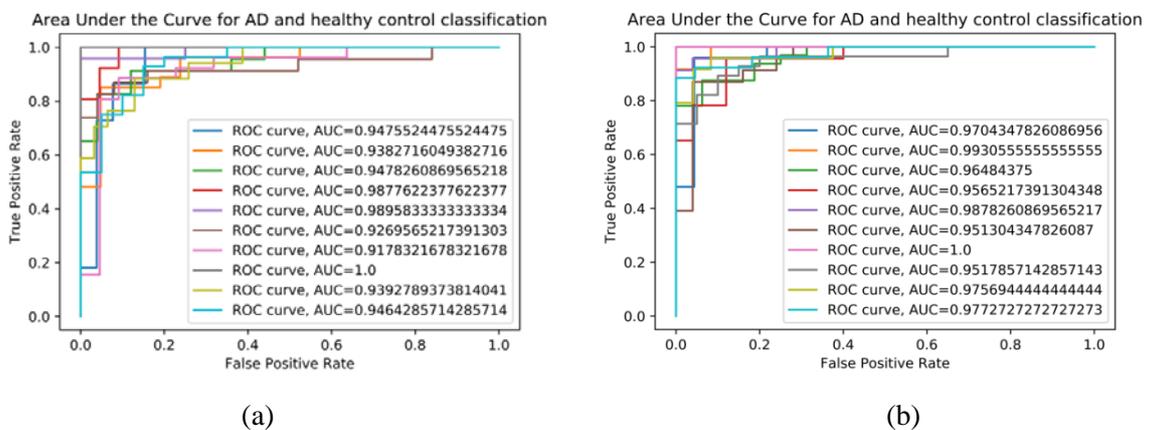

(a)                                         (b)

**Figure 5: ROC and AUC for Pipeline 3 (logistic regression + word count + AD-related features + Doc2vec vectors) on testing set & AUC for Pipeline 4 (logistic regression + word count + AD-related features + Doc2vec & ELMo vectors) on testing set**

Given that the neural network may have randomness, I run the experiment for each model multiple times to get a distribution for test accuracy. As the model gets becomes more complex, its accuracy shows a systematic increase. The output histogram is plotted below in Figure 6:

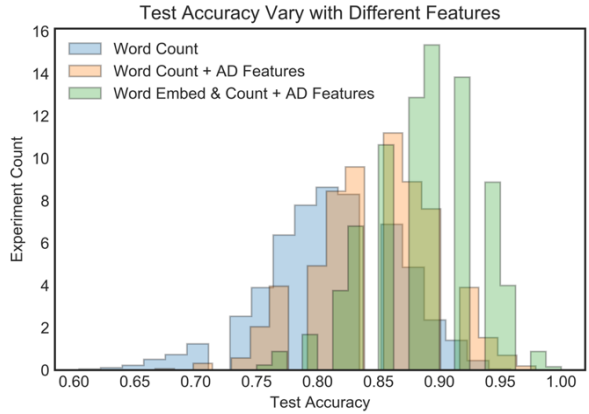

**Figure 6: Test Accuracy Distribution in Three Proposed Models**

Figure 6 shows the standard deviations, ranges, and averages of a total 3000 accuracies generated by the three different pipelines. Each pipeline is set to repeat data splitting and generate accuracy 1000 times. The three bar graphs that indicate the performances of three different pipelines are described below:

- The blue bars represent test accuracy distribution from the first pipeline which only uses the word count feature. From Figure 6, the accuracies generated by the first pipeline range from 0.60 to 0.95 and have an average of 0.81. The wide range of 0.35 indicates that the first pipeline is relatively unstable and also has a relatively low mean accuracy.

- The orange bars represent test accuracy distribution from the second pipeline. This pipeline uses word count and AD-related features such as demographic features and linguistic characteristics. From Figure 6, the accuracies generated by the second pipeline range from 0.68 to 0.98 and have a higher average of 0.86 compared to the first pipeline. The range of 0.3 is narrower, since the second pipeline's stability increases.

- The green bars represent test accuracy distribution from the third pipeline which uses word count, AD-related features, and vector features generated by Doc2vec. From Figure 6, the accuracies generated by the third pipeline range from 0.75 to 1.00 and have a relatively high average of 0.89. The range of 0.25 is much narrower than that of Pipeline 1.

- This graph does not include the test accuracy distribution from the fourth pipeline, which uses word count, AD-related features, and hybrid word embedding (mixed vector features of Doc2vec and ELMo). After splitting data and generating accuracy 1000 times, Pipeline 4 has the highest average accuracy of 0.91 among the four pipelines. The results of all four pipelines are summarized in Table 3 in the Results session.

Figure 6 shows stability and average accuracy gradually increasing. As the range becomes narrower from 0.35 to 0.25, the average accuracy increases from 0.81 to 0.89.

Test results are summarized in Table 3. Test accuracy and model stability improves as the model becomes more complex.

| Method | Test Accuracy | Test AUC | Repetitions |
|---|---|---|---|
| Method 1: Count Vectorizer + Logistic regression | 0.8129 Std: 0.0539 | 0.8897 Std: 0.0464 | 1000 |
| Method 2: AD Features + Count Vectorizer + Logistic regression | 0.8454 Std: 0.0492 | 0.9136 Std: 0.0407 | |
| Method 3: Doc2Vec + AD Features + Count Vectorizer + Logistic regression | 0.8916 Std: 0.0435 | 0.9575 Std: 0.0295 | |
| Method 4: ELMo + Doc2Vec + AD Features + Count Vectorizer + Logistic regression | 0.9125 Std: 0.0403 | 0.9728 Std: 0.0174 | |

**Table 3: Summary Statistics for Test Accuracy and Model Stability using Logistic Regression**

| Model | Accuracy | AUC |
|---|---|---|
| Sarawgi, Utkarsh, et al. [2] | 0.88 | - |
| Fraser et al. [5] | 0.82 | - |
| Mine | 0.9125 | 0.9728 |

**Table 4: Related work model accuracy comparison**

## 5.7. Model Stability

I evaluated model stability by examining standard deviation of test accuracy and test AUC. I find standard deviations of 5.39%, 4.92%, 4.35%, and 4.03% of the test accuracies in four pipelines. The standard deviations of the test accuracies of the four pipelines have a slight decreasing trend as I add more features. The best model has a test accuracy of 91.25% and a standard deviation of 4.03%, which indicates that in terms of accuracy, the model is stable.

## 6. APP Design

To make the early diagnosis accessible to the general public, *AD Scanner*, an online application prototype that incorporates my model was developed. By performing early diagnosis of the fatal dementia Alzheimer's Disease, my app provides a chance for elders to suffer less from AD and alleviate their families' financial burden. Elders also get an accurate diagnosis through only a simple task - speaking for 1 minute about a picture. Currently the working prototype was developed and deployed on-line.

*AD Scanner* is intended to be widely-accessible, efficient, and free as an online application. This overcomes many limitations of conventional medical tests. Firstly, with my app, everybody in need of a pre-screening of AD can easily accomplish the mental evaluation test online. They don't have to frequently visit hospitals to take their medical tests in a long time span. Traditionally, doctors who diagnose AD patients may need various assessments to determine the result, and each visit costs time and money. For elders, especially elders with disabilities, commuting to the hospital is challenging. For example, when an older-aged person in a wheelchair needs to go to the hospital, their family has to lift the wheelchair into the back of the car and carry them in and out. A commute like this three times every month to the hospital is inconvenient. On top of that, receiving test results of every visit can takes several days. However, using my online application at home can save time and significantly reduce financial burden, as it requires no commute at all. It typically only takes around five minutes to finish the screening process and get the result. Secondly, the accuracy of my model is higher than manual mental tests such as MMSE (87%) [19]. My model is accurate and reliable as it has an accuracy of 91% $\pm$ 4.03%. It helps achieve an accurate early diagnosis of AD.

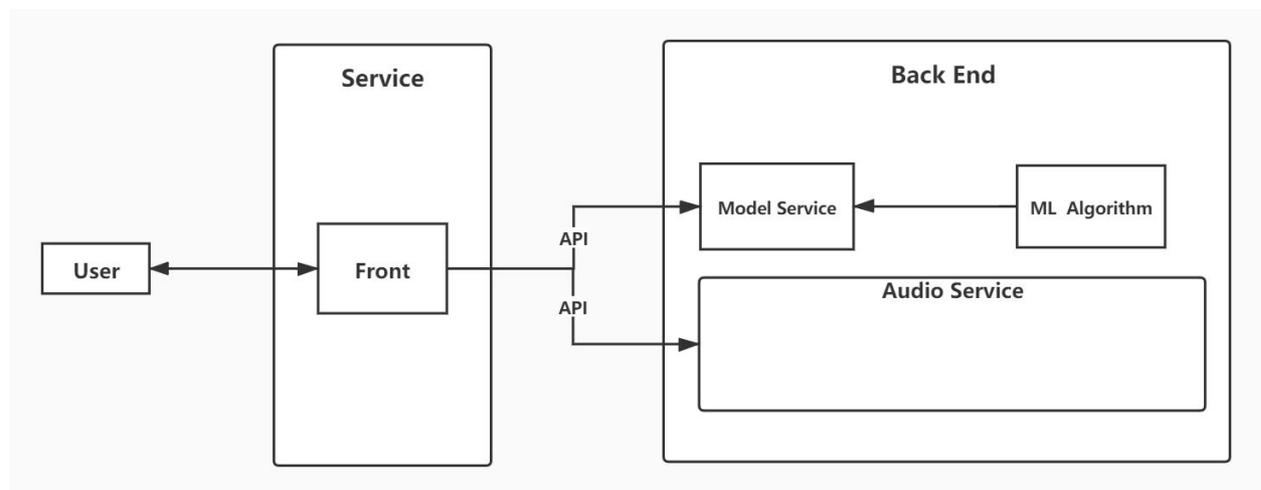

**Figure 7. Online application user interface**

In the online application, the front is constructed by HTML, CSS, and JavaScript code. After receiving a user request, the application turns to Django and starts the requested process in the user interface HTTP. By processing using Views.py, the process response is returned to WSGI. Finally, the response is shown on the user interface.

The second page of the application shows the Boston Cookie Theft picture and a question about the scenario. It asks the user to perform one language task to generate the diagnosis – the Boston Cookie Theft picture description. Below the picture there is a button the user can click to collect their description of the picture.

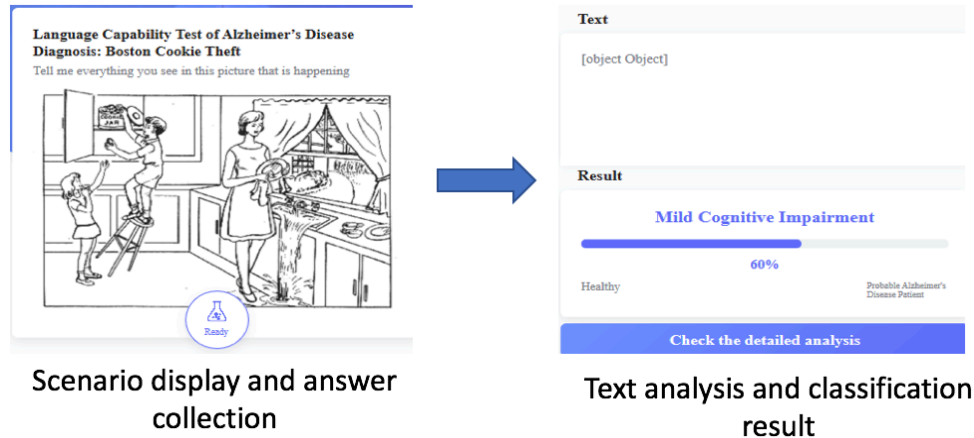

**Figure 8. Online application user interface**

The language task is the same protocol the researchers at the University of Pittsburgh used to collect both the control and dementia data in DementiaBank. This ensures that I can use the model trained on the DementiaBank dataset to evaluate the user's mental status based on the current input answer.

This application can be used as a large screening method for AD specialists. For instance, I offered the service of this application to multiple nursing homes in the US as a field test. I hope to get feedback from this practical scenario.

In the future, this application could turn into a cloud service. By collecting users' data in different languages or with different mental statuses, I hope to improve the accuracy of the model.

## 7. Discussion and Future Work

### 7.1. Performance Boost by Hyperparameter Tuning

Pipeline 3 out of 4 was the first pipeline to incorporate vector features generated by the word embedding algorithm (Doc2vec). Pipeline 4 incorporated hybrid word embedding (Doc2vec and ELMo) features. The initial performance of Pipeline 3 with logistic regression as its classifier, statistical features, demographic features, linguistic characteristics, and Doc2vec features had an accuracy around 0.82. This was even lower than the accuracy of Pipeline 2 (0.84). Recognizing the lack of hyperparameter tuning on Doc2vec, I experimented with hyperparameters vec_size (vector size) and alpha & min_alpha (learning rate).

Vector size decides what dimension the vectors are on. The bigger the vector size, the more specific word embedding for each word and sentence. For example, a 10-dimension vector describes a much more specific location than a 3-dimension vector does. Consider a 1-dimension vector. The location will be a dot in the universe, which is never easy to find. Larger vector size provides a much more specific position of the sentence and word in the context of the paragraph and the document. The initial vector size was 40, which was too small to be accurate. After experimenting with possible vector size values ranging from 40 to 1500 (recorded in Table 2 in session 5.1. Hyperparameter Tuning), the combination of vector size 50, 0.25 for alpha, and 0.0025 for min_alpha yielded an accuracy of 0.89 (alpha and min_alpha is discussed in paragraph after next paragraph).

There are two possible explanations on why a vector size as large as 1500 does not work better than 500. Accuracy peaks and then starts to decline when vector sizes become too large. Firstly, vectors of 1500 dimensions overload the classifier. The model becomes extremely slow, losing one of its efficiency advantages. Secondly and more importantly, the accuracy of the model becomes unreliable when paired with extremely specific features of the training set. The model becomes over-trained on classifying the training data. Then, it will be of no value paired with other pieces of data and may stop predicting completely.

### 7.2. Practical Implications

Manually diagnosing AD may involve too many steps for the average individual: clinic visits, verbal tests, or even laboratory tests. However, even with these complicated procedures, identifying AD is still critical. Language deficit is one of the most noticeable and common features of dementia patients. This, combined with NLP's exceptional development with deep learning applications, inspired a model that uses the NLP algorithm to retrieve information from patient interview data and predict the likeliness of future AD. The model also performs novel grammar pattern detection and pause detection to improve the accuracy of early diagnosis.

### 7.3. Future Work

The application I developed is a prototype and needs further improvement. I am considering adding improvements on different perspectives.

First, the capability to analyze multi-language inputs is my next step. In this way, the model and application can be even more useful to patients worldwide. Languages like Mandarin, Spanish, and German, etc. can be detected once I develop the multi-language application. I am also considering the detection of languages with accents, like Taiwanese. My dataset, DementiaBank, has data for all the languages mentioned [14].

Secondly, I could improve the scalability of the application to serve more users. The process of analyzing the language input on the app is a heavy task. The service of the application will be much quicker after the reduction of platform runtime. I am considering using the Kubernetes container runtime. It is a Docker engine that can scale up or down based on user needs.

## 8. Conclusion

In this paper, I show that incorporating NLP elements in logistic regression can significantly improve classification accuracy. I also improve accuracy and AUC by enhancing the model with hybrid word embedding algorithms. Introducing a feature embedding algorithm not only boosts accuracy but also increases model's stability. Although future improvement is possible, my current result will provide medical professionals with a cheaper and simpler way to early diagnose AD.

My model focuses on providing classification accuracy in three distinct ways. First, I incorporate NLP elements in logistic regression. Second, I enhance my model with hybrid word embedding algorithms, which also improve AUC. Third, I introduce a feature embedding algorithm that also increases model stability. Although there are future improvements to be made, my current results will provide medical professionals with a simpler and inexpensive way to diagnose early AD.